\documentclass[]{heygen_research}

\usepackage[toc,page,header]{appendix}
\usepackage{natbib}

\usepackage{amsmath}
\usepackage{amssymb}
\usepackage{dsfont}

\usepackage{cleveref}

\usepackage{subcaption}

\usepackage{xargs}
\usepackage{todonotes}

\usepackage{multirow}

\usepackage{algorithm}
\usepackage{algorithmic}

\usepackage{array}
\usepackage{tabularx}

\usepackage{tcolorbox}

\usepackage{adjustbox}

\usepackage{bm}
\usepackage{mathrsfs}

\usepackage{adjustbox}
\usepackage{multicol}
\usepackage{changepage}
\usepackage{graphicx}
\usepackage{array}
\usepackage{hyperref}

\usepackage{minitoc}

\newcommand{\eg}{\emph{e.g., }}

\newcommand{\model}{Avatar~V}

\title{Avatar V: Scaling Video-Reference Avatar Video Generation}

\author[]{HeyGen Research}

\abstract{
Generating avatar videos that are not merely visually similar to a target individual but behaviorally recognizable, faithfully reproducing their talking rhythm, gestural tendencies, and expression dynamics, remains an open challenge. Existing methods predominantly condition on single static images, which provide insufficient identity information and cannot capture dynamic motion traits, while standard pixel-level training objectives underserve the perceptually critical facial regions that determine avatar fidelity. To solve these issues, we present \model{}, a production-scale framework that addresses these limitations through video-reference-conditioned identity modeling. Rather than compressing identity into fixed-size embeddings, the model conditions directly on the full token sequence of a reference video, learning to extract and reproduce both static identity attributes (facial geometry, skin texture) and dynamic behavioral patterns (talking rhythm, micro-expressions) through attention over the reference context. To make this formulation practical, we introduce \emph{Sparse Reference Attention}, an asymmetric mechanism that achieves linear-complexity conditioning on arbitrarily long references. To capture individual-specific motion style, we propose a dedicated motion representation stream that enables \emph{closed-loop talking style transfer}. To recover perceptually critical facial details at production resolution, we design an \emph{identity-aware super-resolution refiner} that inherits the full reference conditioning apparatus. These components are supported by a scalable data engine curating 100M+ training clips from 50M raw videos with cross-clip identity connectivity, and a five-stage progressive training pipeline incorporating flow matching pre-training, personality fine-tuning, two-phase distillation ($>$10$\times$ acceleration), and RLHF alignment, deployed across thousands of GPUs. \model{} generates 1080p videos of unlimited duration, achieving state-of-the-art performance across identity preservation, lip synchronization, and generation quality on our cross-scene benchmark, consistently outperforming leading systems including Seedance 2.0, Kling O3 Pro, Veo 3.1, and OmniHuman 1.5 in both automated metrics and human evaluation.
}

\checkdata[Project Page]{\url{https://www.heygen.com/research/avatar-v-model}}

\begin{document}

\maketitle

\newpage
\tableofcontents
\newpage

% ============================================================
\section{Introduction}
\label{sec:intro}

Over the past year, the field of video generation has undergone a decisive shift from unimodal synthesis toward multimodal, controllable, and identity-aware generation. Proprietary systems such as Sora~\citep{brooks2024sora}, Kling~\citep{kling2024}, and Seedance~\citep{seedance2025,seedance15pro2025}, alongside open-source models including Wan~\citep{wan2025}, CogVideoX~\citep{cogvideox2024}, and HunyuanVideo~\citep{hunyuanvideo2024}, have transformed video generation from a research curiosity into a practical, utility-driven capability. In parallel, audio-driven portrait animation has seen rapid progress through diffusion-based approaches~\citep{xu2024hallo,cui2025hallo3,cui2025hallo4,tu2025stableavatar,meng2026echomimicv3,gan2025omniavatar,chen2025humo}, enabling increasingly realistic talking-head synthesis from reference images and driving audio signals.

Despite this remarkable progress, generating production-quality talking avatar videos, where a digital human faithfully reproduces a real person's appearance, expressions, and talking style across diverse scenes, remains an open and multi-faceted challenge. Current systems still fall short in three fundamental aspects:

\begin{itemize}[leftmargin=*]

\item \textbf{Shallow identity representation.} Nearly all existing methods condition generation on a single static reference image~\citep{xu2024hallo,meng2026echomimicv3,gan2025omniavatar}, which captures the subject from one viewpoint, under one lighting condition, with one expression. This forces the model to hallucinate unseen views and articulation patterns, leading to identity drift, loss of fine-grained facial details, and an inability to reproduce the individual's characteristic talking style. Recent works explore video-based references~\citep{cheng2025wananimate,lai2026slotid,seedance2_02025}, but naively concatenating all reference tokens with generation tokens incurs prohibitive quadratic attention cost. Moreover, these approaches lack explicit supervision on both static identity similarity (facial geometry, skin texture) and dynamic motion fidelity (talking rhythm, expression dynamics).

\item \textbf{Decoupled appearance and motion style.} A convincing avatar must not only \emph{look like} the target person but also \emph{move like} them. Existing systems typically treat identity as a static embedding and motion as a separate conditioning signal, failing to capture the individual's talking rhythm, habitual micro-expressions, and gestural tendencies.

\item \textbf{Sparse supervision for perceptually critical regions.} Standard diffusion training optimizes a pixel-level loss that distributes learning signal uniformly across the frame, yet the regions most critical for avatar quality (lip shape, teeth, micro-expressions, eye gaze) occupy a small fraction of total pixels. This leads to undertrained facial details and poor lip synchronization, and conventional training pipelines have not been systematically adapted for identity-preserving avatar generation.

\end{itemize}

To address these challenges, we present \model{}, a reference-video-based personalized large video generation model for production-scale talking avatar synthesis. The central idea is to formulate personality embedding as a \emph{video-reference conditioning} problem: rather than compressing identity into fixed-size embeddings, the model conditions directly on the full token sequence of the user's reference video, learning to extract and reproduce both static identity attributes and dynamic behavioral patterns through attention over the reference context. Given a short reference video of any individual, \model{} generates 1080p avatar videos of infinite duration that faithfully preserve the target person's appearance and talking style. \model{} has been deployed to serve millions of generation requests. Our contributions span model architecture, data curation, and training methodology:

\begin{itemize}[leftmargin=*]

\item \textbf{Video-reference identity conditioning via Sparse Reference Attention.} We condition generation on full video references of arbitrary length through an asymmetric attention mechanism that models both static identity features (facial geometry, skin texture, accessories) and dynamic behavioral patterns (talking rhythm, habitual expressions, gestural tendencies). Generation tokens attend to all reference tokens for fine-grained identity extraction, while reference tokens only self-attend, reducing complexity from quadratic to linear in reference length (Section~\ref{sec:model}).

\item \textbf{Talking style modeling via motion representation.} We introduce a dedicated motion stream that simultaneously serves as a generation target and a conditioning signal, creating a closed-loop training signal for learning each individual's characteristic motion patterns. Through joint optimization of these dual roles, the model develops a unified understanding of the target speaker's motion dynamics (Section~\ref{sec:model}).

\item \textbf{Identity-aware super-resolution refiner.} We design a super-resolution module that inherits the full video reference conditioning apparatus, leveraging identity and audio signals to recover fine-grained facial details lost at base resolution, with efficient sparse temporal attention for practical high-resolution inference (Section~\ref{sec:model}).

\end{itemize}

\noindent These architectural contributions are supported by a co-designed data and training infrastructure:

\begin{itemize}[leftmargin=*]

\item \textbf{Scalable data curation with cross-identity connectivity.} We build a data engine producing training data at three quality tiers from over 50M raw videos, with an identity-aware cross-clip connectivity graph that links same-identity clips across visually distinct scenes for disentangling identity from scene-specific details (Section~\ref{sec:data}).

\item \textbf{Human-aware progressive training.} We develop auxiliary losses in learned representation spaces (identity, motion, lip-sync, perceptual fidelity) integrated into a five-stage pipeline spanning text-to-video pretraining, audio-to-video pretraining, personality SFT, two-phase distillation for over $10\times$ acceleration, and reinforcement learning from human feedback (Section~\ref{sec:training}).

\item \textbf{End-to-end production system.} We deploy the full pipeline across 5{,}000+ GPUs with inference optimizations including context caching, sequence parallelism, fused kernels, and streaming VAE decode, under a unified multi-cloud infrastructure with QoS-aware scheduling (Sections~\ref{sec:inference} and~\ref{sec:infra}).

\end{itemize}

Comprehensive experiments on our cross-scene benchmark demonstrate state-of-the-art performance across all evaluated dimensions, consistently outperforming leading systems including Seedance 2.0, Kling O3 Pro, Veo 3.1, and OmniHuman 1.5 in both automated metrics and human evaluation.

% ============================================================
\section{Model Design}
\label{sec:model}

\begin{figure*}[!ht]
  \centering
  \includegraphics[width=\textwidth]{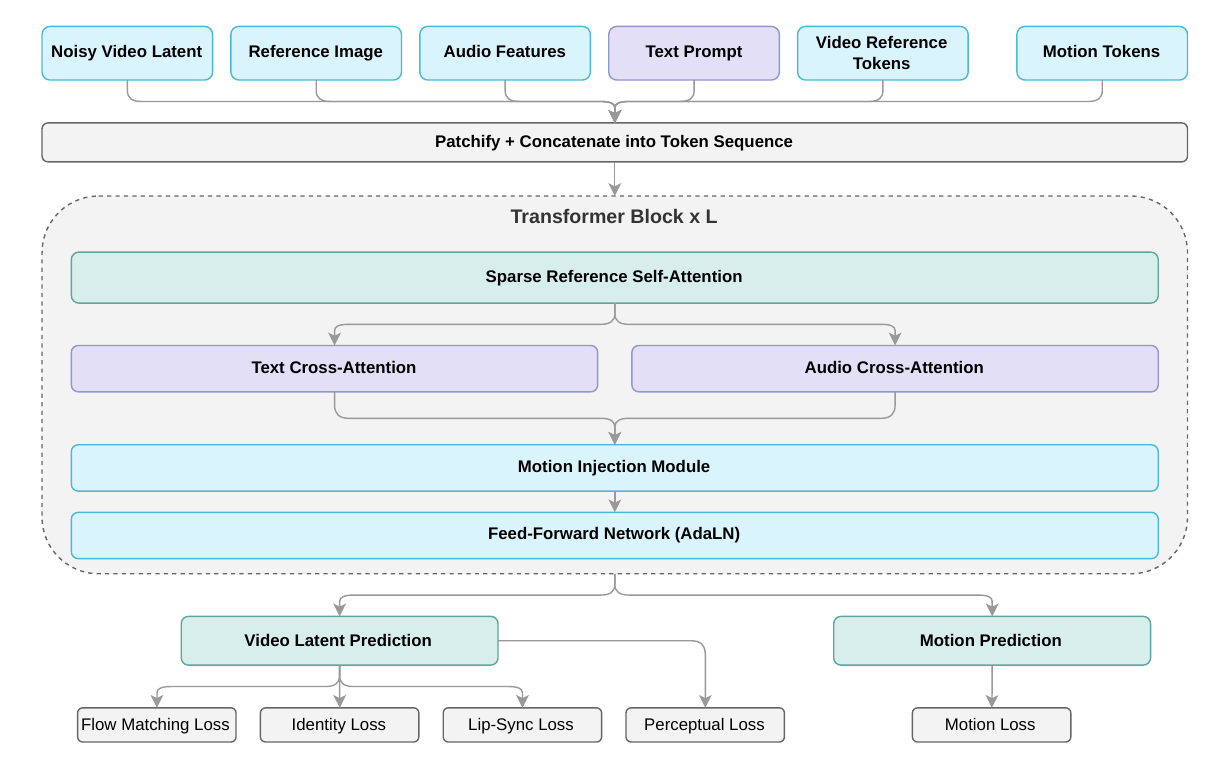}
  \caption{\textbf{\model{} Architecture.} Multi-modal inputs are patchified into a unified token sequence and processed through $L$ transformer blocks. Each block contains Sparse Reference Self-Attention, text and audio cross-attention, a Motion Injection Module, and an AdaLN-modulated feed-forward network. The model produces a video latent prediction supervised by flow matching and human-aware losses, alongside an auxiliary motion prediction.}
  \label{fig:architecture}
\end{figure*}

We present \model{}, an identity-preserving, audio-driven video generation system built on a Diffusion Transformer (DiT) architecture with the flow matching training framework. \model{} takes as input a short user video (ranging from a few seconds to several minutes), a target audio track, and a text prompt describing the desired scene, and generates high-fidelity high resolution avatar videos that faithfully reproduce the target person's appearance, expressions, and talking style. The system comprises four major components: an Identity-Preserving Image Engine for scene image generation, a core DiT model with Sparse Reference Attention for personality-embedded video generation, an identity-aware super-resolution refiner for high-resolution output, and a curated data pipeline that enables cross-identity training. The overall architecture is illustrated in Figure~\ref{fig:architecture}.

\subsection{VideoRef DiT: Video-Reference Personality Modeling}

The core of \model{} is a Diffusion Transformer that formulates personality embedding as a \emph{video-reference conditioning} problem: rather than compressing identity into low-dimensional embeddings or fixed-size feature vectors, the model directly conditions on the full token sequence of the user's reference video, learning to extract and reproduce fine-grained identity details through attention at every transformer layer. The reference video tokens serve as a rich identity context: the model observes the target person's appearance, expressions, and motion patterns, and generates new video that is consistent with these observations. This approach offers key advantages over parametric identity encoding: it scales naturally with reference length (more context yields richer identity information), requires no identity-specific fine-tuning at inference time, and preserves the full visual richness of the reference rather than discarding information through a bottleneck.

\paragraph{Design objectives}
The architecture is guided by three objectives that are jointly addressed through the components described below. \emph{Long-form temporal consistency}: the model must maintain stable identity, coherent motion, and consistent scene composition across sequences ranging from several seconds to over a minute. \emph{Audio-visual synchronization}: dedicated audio cross-attention modules (Section~\ref{sec:model}) align speech content with visual articulation at the phoneme level. \emph{Natural motion dynamics}: beyond lip movements, the model must reproduce co-speech gestures, gaze shifts, and postural sway that collectively determine perceived naturalness.

\paragraph{Static and Dynamic Identity Modeling}
What distinguishes \model{} from existing systems is its ability to model both \emph{static} and \emph{dynamic} aspects of personal identity. Static features include fine-grained, time-invariant characteristics such as dental structure, skin texture and wrinkles, facial geometry, hair style and color, and accessories. Dynamic features encompass the individual's characteristic behavioral patterns: talking rhythm and mouth movement amplitude, habitual micro-expressions and smile characteristics, and gestural tendencies during speech. The model supports reference videos of arbitrary length: short references provide basic appearance information, while longer references enable the model to observe and internalize the individual's talking cadence and expression dynamics. This scalability is achieved without architectural modification, allowing the system to gracefully adapt to the available reference material. The result is that generated videos are not merely facially similar to the target but are \emph{behaviorally recognizable}: the generated person looks like and acts like the target individual.

\paragraph{Sparse Reference Attention}
To fully exploit the dynamic visual features contained in the user's reference video while maintaining computational tractability, we introduce Sparse Reference Attention, a structured sparsity mechanism for video-reference identity conditioning. Standard approaches either compress references into low-dimensional bottlenecks that lose fine-grained identity details, or concatenate all reference tokens with generation tokens incurring prohibitive quadratic cost as reference length grows. Sparse Reference Attention addresses this trade-off through a carefully designed sparsity pattern that preserves full access to identity information during generation while eliminating redundant computation among tokens that do not require mutual interaction. The resulting complexity scales almost linearly with reference length, enabling the model to condition on minutes-long reference footage that captures not only static appearance but also the subject's characteristic expressions, gestures, and talking rhythm, without architectural modification.

\paragraph{Talking Style Modeling via Motion Representation}
Talking style, the characteristic temporal pattern of facial movements, mouth shapes, and head gestures during speech, is a crucial but challenging aspect of identity. We observe that talking style can be understood as a temporal variation pattern over motion representations: the same phoneme sequence produces visually different articulation patterns depending on the speaker's individual style. To capture this, we introduce a dedicated motion representation stream that serves as both a learning objective and a conditioning signal within the model. Through joint optimization of these two roles, the model develops a unified understanding of the target speaker's motion dynamics, enabling it to both internalize and reproduce characteristic talking patterns. This integrated design yields faithful style transfer even for unseen speech content, producing generated videos that are behaviorally consistent with the reference speaker.

\paragraph{Human-Aware Auxiliary Losses}
Traditional audio-to-video training relies solely on pixel-level diffusion loss, which provides insufficient learning signal for subtle but perceptually critical features like talking style and micro-expressions, which represent a small fraction of total pixel variation yet are essential for identity perception. To address this, we introduce a suite of human-aware auxiliary losses that provide semantically meaningful supervision beyond raw pixels, covering identity consistency, motion fidelity, audio-visual synchronization, and perceptual quality. These losses operate in learned representation spaces rather than pixel space, providing denser and more informative supervision for the human-centric aspects of avatar generation.

\subsection{Identity-Preserving Image Engine}

The \model{} pipeline begins with the construction of a high-fidelity, identity-preserving scene image that serves as the visual anchor for subsequent video generation. Given only a short user-provided video, the Image Engine is tasked with generating a photorealistic image of the user in a novel scene while faithfully preserving their unique facial identity. A key design principle is the efficient and thorough exploitation of the user's input video: rather than relying on a single reference frame (which may suffer from suboptimal pose, expression, or occlusion), the pipeline automatically selects a diverse set of frames spanning multiple viewpoints and expressions. This multi-view, multi-expression sampling strategy ensures that the identity representation is both comprehensive and robust, supplying the generation engine with sufficient information to hallucinate consistent identity across novel viewpoints while reproducing subtle identity cues such as smile asymmetry, dimple patterns, and nasolabial fold characteristics. The resulting scene image satisfies several quality criteria: identity fidelity in a learned embedding space, scene diversity through text-prompt-controlled backgrounds and lighting, photorealism with natural skin texture and coherent illumination, and flexible resolution and aspect ratio to accommodate diverse downstream video generation requirements.

\subsection{Audio Engine: LLM-Based Voice Cloning}

In addition to visual identity, faithful voice reproduction is essential for convincing avatar videos. The \model{} Audio Engine is a proprietary voice cloning system built on a large language model (LLM) backbone that generates target-speaker speech from arbitrary text input. Given a short audio sample from the user's reference video (as little as 10 seconds), the system extracts a speaker embedding that captures the individual's vocal timbre, prosody patterns, speaking rate, and accent characteristics. The LLM-based architecture models speech generation as a sequence prediction task over discrete audio tokens, enabling it to synthesize natural, expressive speech that faithfully preserves the speaker's voice identity while supporting multilingual output and emotion control. The Audio Engine operates as a standalone module in the \model{} pipeline: for text-to-avatar use cases, it generates the driving audio track from the user's script before passing it to the VideoRef DiT; for audio-driven use cases, the user-provided audio is used directly. This modular design decouples voice synthesis from video generation, allowing each component to be independently improved and scaled.

\subsection{Super-Resolution Refiner with Identity Modeling}

While the base DiT operates at low resolution to maintain computational tractability, production deployment demands high-resolution output. The \model{} Super-Resolution Refiner bridges this gap through an identity-aware upsampling module that enhances visual fidelity without compromising identity consistency. The Refiner shares the same DiT backbone as the base model but accepts the low-resolution output as additional conditioning input alongside the high-resolution noise, enabling it to leverage the base model's predictions as a strong prior for detail synthesis.

\paragraph{Identity-Aware Conditioning}
A naive super-resolution approach would treat upsampling as a purely visual enhancement task, potentially introducing identity-inconsistent artifacts in facial regions. The \model{} Refiner instead inherits the full identity modeling apparatus from the base DiT, including video reference conditioning, audio features, and motion representations, ensuring that facial identity is preserved with high fidelity throughout the upsampling process.

\paragraph{Efficient Inference via Sparse Temporal Attention}
Since the base model has already established strong temporal coherence at low resolution, the Refiner's primary role is local detail enhancement rather than global temporal reasoning. We therefore employ sparse temporal attention that restricts each frame's receptive field to a local neighborhood, significantly reducing the computational cost at high resolution. Through a multi-stage distillation process, the Refiner achieves high-quality upsampling in very few denoising steps, enabling practical inference latency at high resolution.

The model architecture described above places stringent requirements on training data: the Sparse Reference Attention mechanism requires same-identity video pairs across diverse scenes, the motion representation stream demands dense temporal annotations, and the human-aware auxiliary losses rely on per-frame quality signals. The following section describes the scalable data curation pipeline that meets these requirements.

% ============================================================
\section{Training Strategy}
\label{sec:training}

\begin{figure*}[!ht]
  \centering
  \includegraphics[width=\textwidth]{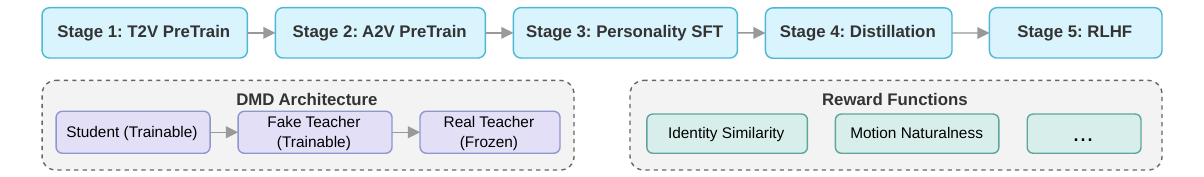}
  \caption{\textbf{\model{} Training Pipeline.} The five-stage curriculum progressively builds capabilities from general video generation through audio-driven synthesis and identity-preserving personality embedding to distillation and human feedback alignment. Bottom panels detail the DMD distillation architecture and GRPO reward functions.}
  \label{fig:training_pipeline}
\end{figure*}

We adopt a progressive multi-stage training paradigm that systematically develops the model's capabilities from general video understanding to identity-specific personality embedding and human-preference alignment. Our training pipeline consists of five major phases: text-to-video general pre-training, audio-to-video pre-training, personality supervised fine-tuning (SFT), distillation, and reinforcement learning from human feedback (RLHF). This structured approach enables the model to learn temporal dynamics, identity preservation, efficient inference, and perceptual quality in a stable and efficient manner. Figure~\ref{fig:training_pipeline} summarizes our training pipeline.

\subsection{Text-to-Video General Pre-Training}

The first stage establishes the model's foundational video generation capabilities through large-scale text-to-video (T2V) pre-training on the diverse pretraining corpus described in Section~\ref{sec:data}. This stage trains only the base-resolution DiT; the super-resolution refiner is not involved.

\paragraph{Progressive resolution and duration scaling}
Following the curriculum strategy adopted by recent video generation models~\citep{seedance2025,cogvideox2024,hunyuanvideo2024}, we progressively increase spatial resolution and temporal length throughout pre-training. Training begins at low resolution with short clips to allow the model to rapidly converge on basic motion patterns and scene composition, then gradually scales to higher resolution and longer sequences. This coarse-to-fine strategy improves both training stability and final generation quality compared to directly training at the target resolution.

\paragraph{Multi-task joint training}
The T2V pre-training phase jointly trains on text-to-video and image-to-video (I2V) generation tasks within a unified framework. For I2V, the first frame is provided as a conditioning image, and the model learns to generate temporally coherent video continuations. This multi-task formulation enables the model to learn both unconditional video dynamics from T2V and frame-conditioned temporal extension from I2V, providing a versatile foundation for downstream avatar-specific fine-tuning.

\paragraph{Training objective}
We adopt rectified flow matching~\citep{lipman2022flow,esser2024scaling} as the training objective, where the model learns to predict the velocity field that transports noise to data along straight-line trajectories. A logit-normal timestep distribution concentrates training signal on intermediate noise levels where the learning signal is most informative. We use a distributed Muon optimizer~\citep{jordan2024muon} for 2D+ weight tensors for improved convergence on large-scale models, and AdamW for embeddings and 1D parameters. A cosine learning rate schedule with linear warmup is applied throughout.

\subsection{Audio-to-Video Pre-Training}

Building on the T2V foundation, the second stage specializes the model for audio-conditioned avatar video generation, following the line of audio-driven portrait animation works~\citep{xu2024hallo,cui2025hallo3,tu2025stableavatar,meng2026echomimicv3}. Starting from the pre-trained T2V checkpoint, the model is adapted to accept a conditioning image (the first frame) and a driving audio track, learning to generate temporally coherent video continuations with synchronized lip movements and natural head motion. This stage introduces the audio cross-attention modules and trains them jointly with the visual backbone on a broad corpus of talking-head video data spanning diverse speakers, languages, and speaking styles. Progressive resolution scaling and dynamic sequence length sampling build robust multi-scale generation capabilities.

\subsection{Personality Supervised Fine-Tuning}

The supervised fine-tuning stage transforms the general-purpose video model into an identity-aware avatar generator by training on the curated same-identity-different-scene dataset described in Section~\ref{sec:data}. During SFT, the model receives reference videos of the target identity through the Sparse Reference Attention mechanism, teaching the model to extract and utilize identity information from video references. The motion representation pathways are activated during this stage, enabling the model to learn talking style transfer. Leveraging the cross-clip connectivity from the data pipeline, each training example consists of a target video clip paired with reference clips from different scenes but the same identity, forcing the model to extract identity-invariant features rather than simply copying scene-specific details. The human-aware auxiliary loss suite is progressively activated, providing dense semantic supervision that guides the model toward identity-faithful, expressively accurate, and well-synchronized generation.

\subsection{Model Distillation}

To enable practical deployment with low inference latency, we apply a two-phase distillation strategy~\citep{salimans2022progressive} that reduces both the number of classifier-free guidance (CFG) evaluations and the number of denoising steps required for generation.

\paragraph{CFG Distillation}
The VideoRef DiT employs multiple classifier-free guidance~\citep{ho2022cfg} streams to control different conditioning aspects of the generated video. At inference time, each CFG stream requires a separate forward pass with the corresponding condition dropped, resulting in a multiplicative increase in computational cost. To eliminate this overhead, we distill the multi-stream CFG behavior into a single forward pass~\citep{meng2023distillation}, reducing the per-step cost by a factor proportional to the number of guidance streams while preserving generation quality.

\paragraph{DMD Distillation}
Following CFG distillation, we further reduce the number of denoising steps using an improved Distribution Matching Distillation (DMD) framework~\citep{yin2024dmd,song2023consistency}. Our implementation employs a three-model architecture: a trainable \emph{student} that generates video from pure noise in a single forward pass, a trainable \emph{fake teacher} that models the student's output distribution, and a frozen \emph{real teacher} that provides the target distribution from the original multi-step model. The student learns to match the real teacher's distribution through a combination of distribution matching gradients and progressive distillation objectives, with an optional adversarial loss for sharpening fine-grained details. Our implementation incorporates several stability improvements over vanilla DMD, resulting in more reliable convergence. The combined two-phase distillation pipeline reduces the total inference cost by over an order of magnitude while maintaining generation quality comparable to the original multi-step, multi-CFG model.

\subsection{Human Feedback Alignment}

The final training stage aligns the model with human perceptual preferences through reinforcement learning from human feedback~\citep{black2024rlhfdiffusion}. We employ multiple reward signals that jointly cover identity fidelity, motion naturalness, and visual quality. We adopt Group Relative Policy Optimization (GRPO)~\citep{shao2024grpo,zhang2025flowgrpo,huang2025dancegrpo} as the primary RL algorithm, adapted with a flow-matching-compatible formulation for efficient policy gradient computation within the diffusion framework. KL regularization against the pre-RLHF model prevents quality degradation on previously learned capabilities. As a complementary approach, we also support Direct Preference Optimization (DPO)~\citep{rafailov2023dpo,wallace2024diffusiondpo} training, which learns directly from human-annotated preference pairs without requiring online generation. The preference pairs used for DPO training are collected through the human annotation system described in Section~\ref{sec:data}.

% ============================================================
\section{Inference}
\label{sec:inference}

\begin{figure*}[!ht]
  \centering
  \includegraphics[width=\textwidth]{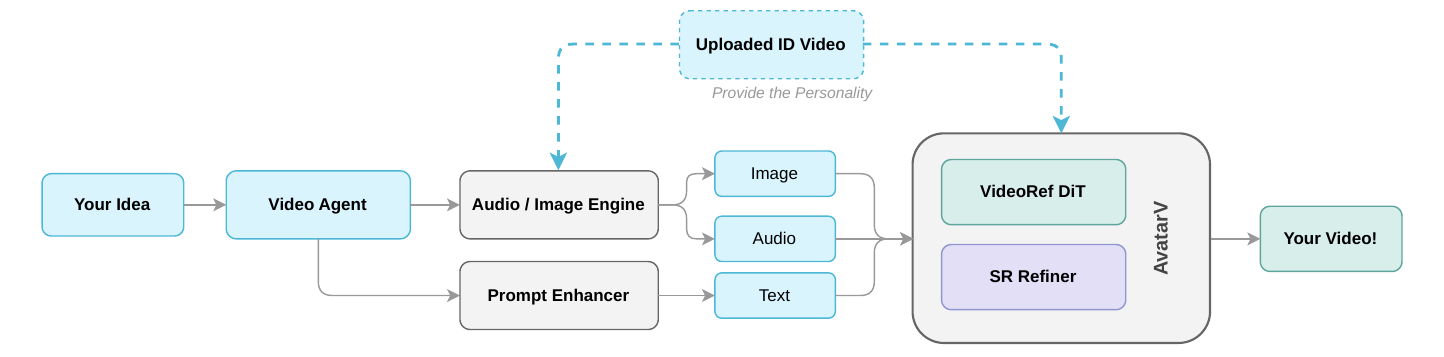}
  \caption{\textbf{\model{} Inference Pipeline.} The user's identity video is processed once into a reusable personality embedding. Scene image generation and prompt engineering proceed in parallel, then all signals are combined for low-resolution DiT generation followed by identity-aware super-resolution to high-resolution.}
  \label{fig:inference_pipeline}
\end{figure*}

This section describes the inference pipeline and the optimizations that enable \model{} to generate high-fidelity, high-resolution talking avatar videos with practical latency. All inference uses the distilled model from Section~\ref{sec:training}, which internalizes classifier-free guidance and operates in a reduced number of denoising steps (24 steps after two-phase distillation), eliminating the need for multiple conditional and unconditional forward passes at each step.

\paragraph{End-to-end pipeline overview}
As illustrated in Figure~\ref{fig:inference_pipeline}, inference proceeds in four stages. (1)~\emph{Preprocessing}: the user's reference video is encoded once into video reference tokens, identity embeddings, and expression embeddings; audio features are extracted from the target audio track; and scene description prompts are encoded into text embeddings. The Identity-Preserving Image Engine generates a scene image conditioned on the reference identity (Section~\ref{sec:model}). These preprocessing steps run in parallel. (2)~\emph{DiT generation}: the base-resolution DiT performs chunk-based autoregressive generation, conditioning on all preprocessed signals through Sparse Reference Attention. (3)~\emph{Super-resolution}: an identity-aware SR refiner upscales the output to high resolution in a single denoising step. (4)~\emph{Streaming decode}: a streaming VAE decoder converts latents to pixels incrementally, producing output frames before the full video is complete.

\subsection{Chunk-Based Long-Form Generation}

\model{} generates videos through a chunk-based autoregressive pipeline that enables arbitrarily long video synthesis while maintaining temporal coherence. Each chunk produces 41 latent frames (corresponding to 161 pixel frames at 25\,fps, approximately 6.4 seconds of video), and chunks are connected through a prefix conditioning mechanism.

\paragraph{Ref2V + Prefix2V Strategy}
The first chunk operates in \emph{ref2v mode}, where the reference video frame is encoded as the identity conditioning signal. Subsequent chunks operate in \emph{prefix2v mode}: the last frames of the previous chunk serve as the prefix condition for the next chunk, providing a smooth temporal bridge. Adjacent chunks share a 2-frame overlap to ensure seamless transitions. This sequential strategy eliminates the need for separate interpolation chunks, simplifying the generation flow while maintaining temporal consistency.

\paragraph{Global Appearance Anchor}
For multi-chunk generation, we extract a global appearance anchor from the first generated chunk. This anchor, combined with motion frame propagation between consecutive chunks, ensures that the subject's identity remains consistent across arbitrarily long videos.

\subsection{Diffusion Sampling}

\paragraph{Improved Stochastic Euler}
Deterministic ODE-based samplers for flow matching models struggle with high-frequency details at reduced step counts, manifesting as unstable hand clarity, blurred dental structures, and temporally inconsistent fine textures. We adopt a stochastic overshoot-and-renoise strategy: at each step, the sample is advanced beyond the target noise level by a controlled overshoot factor, then stochastically renoised back to the correct level with fresh Gaussian noise. This controlled stochasticity improves detail recovery in high-frequency regions and prevents the accumulation of discretization errors, enabling high-quality generation in 24 steps with stable hand, teeth, and facial detail quality comparable to much longer sampling schedules.

\subsection{VideoRef Context Caching and Sparse Attention}

A key observation is that video reference tokens, encoded from clean, non-noisy reference frames, remain invariant across denoising steps. We exploit this through a two-level caching strategy:

\paragraph{Context-Level Caching}
At the first denoising step ($t=0$), the full video reference context (latents, audio features, validity masks, expression and identity embeddings) is computed and cached. For all subsequent steps ($t > 0$), this cached context is reused without recomputation, avoiding the substantial cost of re-encoding the reference video at each of the 24 denoising steps.

\paragraph{Attention-Level KV Caching}
Within each transformer block's reference self-attention layer, the key and value projections of video reference tokens are computed once at the first denoising step and cached in GPU memory. Subsequent steps directly concatenate the cached reference KV tensors with freshly computed generation token KV tensors for the asymmetric attention computation, eliminating redundant linear projections and RoPE computations across all steps.

\paragraph{Sparse Validity Masking}
Video reference sequences can contain invalid tokens (e.g., frames where the face is not visible). We implement sparse attention masks that skip computation for these invalid tokens. The validity masks are precomputed per-rank for the sequence-parallel layout and applied to self-attention, cross-attention, and FFN layers, avoiding wasted computation on non-informative reference tokens.

\subsection{Distributed Inference with Sequence Parallelism}

\model{} employs Ulysses Sequence Parallelism (USP) across 8 GPUs within a single node to distribute the computation of long token sequences. The input sequence, comprising video latents, reference tokens, high-resolution face tokens, and conditioning tokens, is partitioned along the sequence dimension across GPU ranks, with all-to-all communication for attention operations that require cross-rank token interaction.

\paragraph{FSDP2 with CPU Offloading}
Model parameters are sharded across GPUs using FSDP2 with CPU offloading for inactive parameter shards. This frees GPU memory for the large intermediate activations required by the DiT and enables multi-model co-location: multiple model variants can reside on a single machine with rapid switching via CPU-to-GPU loading rather than full reloading from disk. Forward prefetching overlaps the AllGather of the next block's parameters with the current block's computation, hiding communication latency. Processes are pinned to the NUMA node of their assigned GPU to ensure optimal memory bandwidth for these CPU-GPU transfers.

\subsection{Inference Acceleration}

Deploying \model{} at production scale surfaces three latency bottlenecks: (1)~kernel launch overhead and memory bandwidth waste from thousands of small operators per transformer block, (2)~coarse-grained inter-GPU synchronization in sequence-parallel attention, and (3)~hardware-level frequency variance across GPU ranks causing straggler effects at collective boundaries. We address each through a custom compiler with agentic kernel synthesis, NVSHMEM-based sequence parallelism, and targeted system-level tuning, achieving a combined 3$\times$ latency reduction over the unoptimized baseline and 33\% latency reduction over the torch inductor compiler optimized version.

\subsubsection{Custom Compiler with Agentic Kernel Synthesis}

\paragraph{Limitations of existing compilers}
While \texttt{torch.compile} with the Inductor backend provides a general-purpose optimization path, we find it insufficient for production diffusion inference at our scale. First, Inductor's pattern matcher fails to capture many cross-operator fusion opportunities, leaving significant memory bandwidth on the table. Second, the Triton kernels it generates are suboptimal for the specific tensor shapes and access patterns in our model. Third, Inductor handles dynamic shapes and dynamic tensor memory operations poorly, relying on a guard-and-recompile strategy that triggers excessive recompilation. Fourth, on a complex production codebase with many control-flow paths, the compilation overhead itself becomes prohibitive.

\paragraph{Agentic kernel synthesis}
We introduce a compiler workflow that combines human expertise with LLM-based kernel generation. In the first phase, engineers profile the end-to-end forward pass and define fusion scopes, identifying which operator subgraphs should be merged into single kernel launches. In the second phase, an LLM-based agent takes each fusion specification and generates CUDA/Triton kernel candidates through an iterative evolution process.

A key challenge is that kernel-level profiling is inherently noisy due to GPU thermal state, memory allocator behavior, and scheduling variance. To mitigate this, we adopt an evolution island strategy: 2--3 islands run in parallel, each exploring different tiling strategies and memory access patterns across 4 candidates per generation. Fitness is evaluated on both kernel latency and numerical accuracy, and the best candidate is selected across islands after a fixed iteration budget. This avoids overfitting to noisy single-point measurements while keeping the search tractable.

\paragraph{Results}
The agentic workflow produces mega kernels that fuse entire non-attention portions of each transformer block into single kernel launches, eliminating intermediate tensor materializations and minimizing memory round-trips. The compiled forward pass reduces from thousands of small kernels to only Flash Attention calls, cuBLAS GEMMs, and a handful of fused mega kernels. This yields approximately 3$\times$ latency reduction over the unoptimized baseline and 33\% improvement over the \texttt{torch.compile} Inductor backend.

\subsubsection{NVSHMEM-Based Sequence Parallelism}

\model{} distributes attention computation across 8 GPUs via Ulysses Sequence Parallelism, which requires all-to-all communication at every transformer block. The standard approach using NCCL operates at kernel-level granularity: each all-to-all is a monolithic operation that must fully complete before downstream compute can begin. We replace this with NVSHMEM-based communication that exploits NVLink for direct GPU-to-GPU data movement with tile-level dataflow control.

With NVSHMEM, individual data tiles can be sent, received, and synchronized within a single fused kernel, enabling sub-tensor pipelining: the all-to-all scatter, cuBLAS GEMM computation, and all-to-all gather overlap at fine granularity rather than executing sequentially. As each tile arrives from a remote rank, it is immediately available for computation without waiting for the full transfer to complete.

A critical design consideration is SM partitioning between communication and compute. Naively dedicating SMs to communication leaves the compute GEMM with a non-round number of thread block waves, causing up to 1.5$\times$ slowdown from the wave quantization effect, where SMs sit idle during the final partial wave. We determine the optimal partition through extensive profiling-guided auto-tuning that balances NVLink bandwidth utilization against GEMM wave efficiency.

\subsubsection{System-Level Optimization}

\paragraph{NUMA-aware process placement}
Each inference rank is pinned to the CPU cores and memory controllers on the same NUMA node as its assigned GPU, ensuring optimal bandwidth for the CPU-GPU parameter transfers required by FSDP2 offloading.

\paragraph{GPU clock locking}
In distributed inference where all ranks synchronize at collective operations, the slowest rank determines overall latency. Default GPU boost clocking allows frequency to vary across GPUs depending on thermal and power state, creating straggler ranks that gate every synchronization point. We lock all GPUs to a stable frequency below the boost ceiling, eliminating this variance and reducing overall block latency by approximately 3\%.

\subsection{Super-Resolution}

The base DiT generates video at low resolution in latent space. To produce the final high-resolution output, we apply a single-step adversarial SR model that focuses computational resources on high-detail regions, particularly the mouth area for lip-sync fidelity. Low-resolution latents are noised at $\sigma = 0.6$ before the SR step, providing the model sufficient room for detail enhancement while preserving structural content from the base generation.

\subsection{Streaming VAE Decode}

The VAE decoder employs causal 3D convolutions with temporal feature caching, enabling chunk-by-chunk decoding without requiring the full video to be held in memory. Decoded frames are piped directly into an asynchronous streaming video encoder that writes the output file incrementally. This streaming pipeline ensures bounded memory consumption regardless of video length and enables the first frames to be available before the full video has been decoded.

% ============================================================
\section{Data Curation}
\label{sec:data}

Training \model{} requires two distinct data regimes: a large-scale \emph{pretraining} corpus of diverse human-centric video for learning general motion and appearance priors, and a curated \emph{audio-to-video (A2V) fine-tuning} corpus with dense avatar-specific annotations for talking-head generation. Both corpora are produced by a unified distributed pipeline that orchestrates 25+ processing stages and 20+ specialized AI models across heterogeneous CPU and GPU infrastructure managed by our custom-built data processing engine (Section~\ref{sec:infra}). Through this meticulous curation process, we ultimately obtain over 100M clips for pretraining and 10M+ clips for avatar fine-tuning. 

\subsection{Pretraining Data}

The pretraining data is designed to capture general human-centric video priors at scale. The data is processed through a multi-stage pipeline for filtering, annotation, and feature extraction.

\subsubsection{Segment-Level Curation}

Raw segments pass through a 10-stage cascade. \textbf{Normalization} standardizes resolution (longest side 640px) and frame rate (25\,fps), while \textbf{temporal pre-filtering} rejects choppy or static content via frame-difference analysis and perceptual hashing, both CPU-only, eliminating degenerate content before model inference. \textbf{Human detection} uses a joint object detector and face analysis model to verify human presence and define eligible temporal intervals. \textbf{Optical flow} quantifies motion statistics that feed into the clipping optimizer, and \textbf{visual quality assessment} via Q-Align scores keyframes with continuous quality scores calibrated to human opinion.

\textbf{Smart clipping} formulates clip selection as a constrained optimization problem, jointly maximizing clip duration while satisfying constraints on visual quality, motion, and face presence ratio, replacing brittle independent threshold cascades used in prior work. \textbf{Scene-cut detection} and \textbf{content filtering} via VLMs identify scene boundaries and reject screencasts, game footage, and static photo content. Finally, clips are \textbf{categorized} across 15 semantic dimensions for distribution balancing, and \textbf{video embeddings} are extracted for deduplication.

\subsubsection{Deduplication and Feature Extraction}

GPU-accelerated nearest-neighbor indexing over video embeddings groups near-duplicate clips into clusters; within each cluster, only the highest-quality clip is retained. Rule-based derived categories enable distribution rebalancing across content types.

Deduplicated clips are then processed through 13 parallel extraction stages: (1)~\textbf{visual analysis}: OCR text detection, lip-sync scoring, whole-body pose estimation with dense keypoints, anatomical quality scoring, and synthetic audio detection; (2)~\textbf{audio analysis}: language identification, speaker diarization, and ASR with word-level timestamps; (3)~\textbf{captioning and embeddings}: an in-house audio-video understanding captioner producing rich descriptions, plus text embedding pre-encoding for diffusion conditioning; and (4)~\textbf{latent pre-encoding}: multiple video VAE architectures producing latent representations for direct diffusion transformer training.

\subsection{Audio-to-Video Fine-Tuning Data}

The audio-to-video fine-tuning data applies additional avatar-specific curation to produce training data tailored for talking-head and portrait animation. Ten additional fine-grained quality signals are computed per-clip: eye gaze and blink patterns, face clarity, teeth and hand quality, mouth openness, camera shake, choppy frame detection, lighting consistency, and secondary speaker presence. These composable signals enable flexible quality tier construction without re-running inference.

\begin{figure}[!t]
  \centering
  \includegraphics[width=\textwidth]{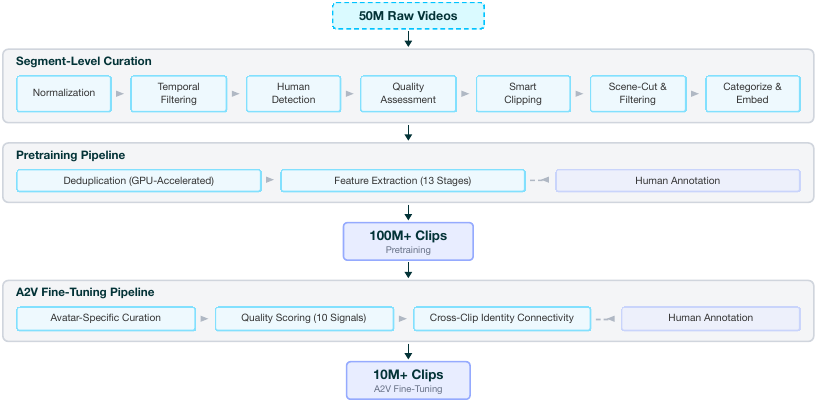}
  \caption{\textbf{Data Curation Pipeline Overview.} Starting from 50M raw videos, our pipeline applies shared segment-level curation before branching into pretraining (100M+ clips) and A2V fine-tuning (10M+ clips) paths, with human annotation and cross-clip identity connectivity feeding into the avatar-specific branch.}
  \label{fig:data_pipeline}
\end{figure}

\subsection{Human Data Annotation System}

Achieving the highest quality thresholds, particularly for RLHF and quality model training, requires reliable human judgment at scale. We build a distributed annotation platform supporting 100+ concurrent freelance annotators across multiple geographic regions.

\textbf{Annotation tasks} span five categories: (1)~\emph{quality scoring} of curated clips across perceptual dimensions (visual quality, facial naturalness, lip-sync accuracy, motion smoothness) using calibrated Likert scales, serving as ground truth for automated quality models; (2)~\emph{preference labeling} via pairwise comparisons of generated outputs along axes of identity preservation, expression naturalness, and audio-visual synchronization, producing training data for DPO and GRPO reward models (Section~\ref{sec:training}); (3)~\emph{bad case filtration} to identify subtle artifacts escaping automated filters, temporal identity drift, teeth deformation, occlusion glitches, asymmetric blinking, with flagged samples feeding back into both corpus cleaning and new detector development; (4)~\emph{generation evaluation and competitive benchmarking} through blinded side-by-side comparisons of model iterations and competitor systems, producing statistically grounded win-rate matrices that guide model improvement priorities and release decisions; and (5)~\emph{attribute annotation} of gaze direction, emotion, gesture type, and speaking style for conditional generation supervision.

\paragraph{Organization} The workforce follows a three-tier hierarchy: \emph{Tier~1} annotators (100+ freelancers) complete qualification pipelines with calibration exercises before admission to production tasks; \emph{Tier~2} reviewers perform random audits on annotation batches and adjudicate disagreements to produce gold-standard labels; \emph{Tier~3} task designers define schemas, write guidelines, and manage the feedback loop between annotation results and model improvement.

\paragraph{Incentive system} Base compensation is supplemented by quality bonuses tied to agreement rates with reviewer audits. A leaderboard tracks accuracy and consistency, with top performers receiving priority access to higher-paying tasks. Annotators falling below thresholds are assigned re-calibration exercises. This closed-loop system maintains inter-annotator agreement above 85\% across all task types.

\subsection{Cross-Clip Identity Connectivity}

Avatar video synthesis demands paired clips depicting the same individual across visually distinct contexts, enabling the model to disentangle identity from background, lighting, and pose. We establish cross-clip connectivity through joint filtering: two clips are linked if they depict the same individual (verified by high face similarity) in visually distinct scenes (verified by low background similarity), with sufficient duration for learning dynamic features. The resulting connectivity graph enables efficient sampling of cross-scene reference pairs during training, organized into resolution-duration groups with balanced demographic representation. This curated data, together with the annotation signals described above, feeds directly into the progressive training pipeline described next.

% ============================================================
\section{Infrastructure}
\label{sec:infra}

Training and data processing for \model{} require coordinating large-scale GPU workloads across heterogeneous, multi-cloud clusters while maintaining high utilization and fault tolerance. This section describes the two core infrastructure systems that support \model{}: \emph{HELIOS}, a unified GPU infrastructure platform for multi-cloud orchestration, and our custom-built \emph{data processing engine} that replaced Ray to handle the scale and scheduling requirements of our video data pipelines.

\subsection{HELIOS: Unified GPU Infrastructure Platform}

As our models and products scaled, training, inference, and data processing increasingly competed for the same scarce GPU resources. The challenge was not merely acquiring more GPUs, but making resources from different providers, regions, and supply models behave as a single usable platform, without requiring every team to understand the underlying complexity.

We built \textbf{HELIOS} (\textbf{H}eyGen \textbf{E}ngine for \textbf{L}arge-scale \textbf{I}nfrastructure \textbf{O}rchestration \textbf{S}ervice), a unified GPU infrastructure platform for multi-cloud and large-scale operations. Today, HELIOS manages more than 5{,}000 GPUs across 5+ providers, 10+ regions, and 15+ standardized cells, supporting reserved, on-demand, and preemptible capacity under one system.

\paragraph{Standardized Onboarding}
Before HELIOS, onboarding a new GPU provider or region required repeating the same engineering work (adapting networking, storage, cluster management, monitoring, and operational workflows) for each new environment. HELIOS replaces this with a standard onboarding path: a new provider or region undergoes common validation, acceptance checks, and baseline infrastructure setup before joining the platform. Once admitted, its resources are exposed through the same management model as the rest of the fleet. This reduced the average time from initial validation to production availability from two weeks to three days.

\paragraph{Cell-Based Architecture}
Rather than building one monolithic cluster, HELIOS organizes the fleet into standardized cells, typically aligned by provider and region. Each cell is a Kubernetes cluster with a validated size boundary and a common operational baseline. This design limits blast radius: a problem in one cell is far less likely to propagate fleet-wide. It also provides a clean growth path by adding capacity through new standard cells rather than stretching a single control plane.

\paragraph{Two-Stage QoS-Aware Scheduling}
Inference workloads require higher priority and faster response; training workloads need larger, more stable allocations over longer periods; data processing is more flexible and can tolerate interruption. Treating all workloads identically would either waste expensive capacity or create contention for critical services. HELIOS employs a two-stage scheduling model: a global scheduler makes capacity decisions based on workload QoS class, GPU type, request size, and supply model, while the selected cell handles local deployment and placement. This improved overall GPU utilization by 15\% and reduced non-productive GPU time by approximately 20\%.

\paragraph{Continuous Resource Governance}
HELIOS continuously monitors key health signals across the fleet, including GPU, PCIe, and NCCL-related conditions. Unhealthy nodes are automatically isolated and routed through operational recovery workflows. The platform also detects long-idle or low-utilization resources by combining signals such as GPU utilization, memory usage, task state, and runtime progress, reclaiming and reallocating capacity according to workload priority.

\paragraph{Unified Observability}
The platform collects signals from infrastructure, clusters, workloads, and applications, applying different sampling and retention strategies depending on the use case. In addition to standard metrics, traces, and logs, HELIOS adds finer-grained network-side observability on key nodes to identify communication bottlenecks and performance jitter in training and inference scenarios. This shared operational view shortens debugging loops, improves capacity planning, and makes cost attribution tractable across teams.

\subsection{Data Processing Engine}

Our video AI data processing pipelines run across multiple GPU clusters, sharing compute resources with online inference serving, where nodes are routinely preempted, rescheduled, and terminated. As data processing demand surged to over 100K concurrent tasks, we encountered fundamental scalability limitations in our initial Ray-based infrastructure and built a purpose-built replacement.

\subsubsection{Design Constraints}

Four constraints define the operating environment for our data processing workloads:

\paragraph{Heterogeneous Pipeline Stages}
A video processing pipeline is a DAG of stages with mixed resource profiles: IO-bound media decode and network transfer, GPU-bound model inference, and CPU-bound processing and encoding. If a GPU stage blocks waiting on an upstream IO stage, or CPU stages over-consume capacity and starve the GPU, utilization collapses. The scheduler must be resource-profile-aware rather than treating every stage as a homogeneous unit of work.

\paragraph{Priority-Based Scheduling}
Latency-critical pipelines feed downstream systems with tight SLAs and require preemptive priority, while background pipelines (\eg batch processing, data enrichment, training data preparation) optimize for throughput. Lower-priority pipelines must yield resources immediately when higher-priority work arrives.

\paragraph{GPU Fragmentation}
The cluster runs multiple GPU types. When a stage requires 4 GPUs but available capacity is scattered as single-GPU slots across different nodes, those GPUs are effectively stranded. The engine must adapt placement strategy to hardware type, workload mix, and real-time capacity.

\paragraph{Constant Node Preemption}
Data processing is colocated with production inference, so a GPU machine running a data pipeline can be reclaimed at any time. The engine must treat node failure as a continuous operating condition, detecting it within seconds rather than minutes, and redistribute work without impacting running pipelines on healthy nodes.

\subsubsection{From Ray to a Custom Engine}

We initially adopted Ray for its actor model, unified Python API, and built-in resource management. As concurrent pipeline counts grew, we built an external scheduler on top of Ray for priority scheduling and pipeline-aware resource allocation. However, when the data processing cluster grew to more than 2k nodes, Ray's Global Control Store (GCS) became the bottleneck: it consumed over 100\,GB RSS and 400\% CPU. Profiling revealed that the GCS could not keep up with message volume at this scale: stale metadata accumulated indefinitely with no eviction, and under sustained load, the GCS suffered process crashes that took down the entire coordination layer.

The root cause was architectural: the GCS is a centralized coordination point where every state change must be broadcast to every node, generating traffic that scales quadratically with cluster size. After extensive stabilization attempts (parameter tuning, cache capping, timeout reduction, telemetry disabling, and metric backend switching), each yielding only marginal improvement, we concluded we had outgrown Ray's architecture and built a purpose-built replacement.

\begin{figure}[!t]
    \centering
    \includegraphics[width=\linewidth]{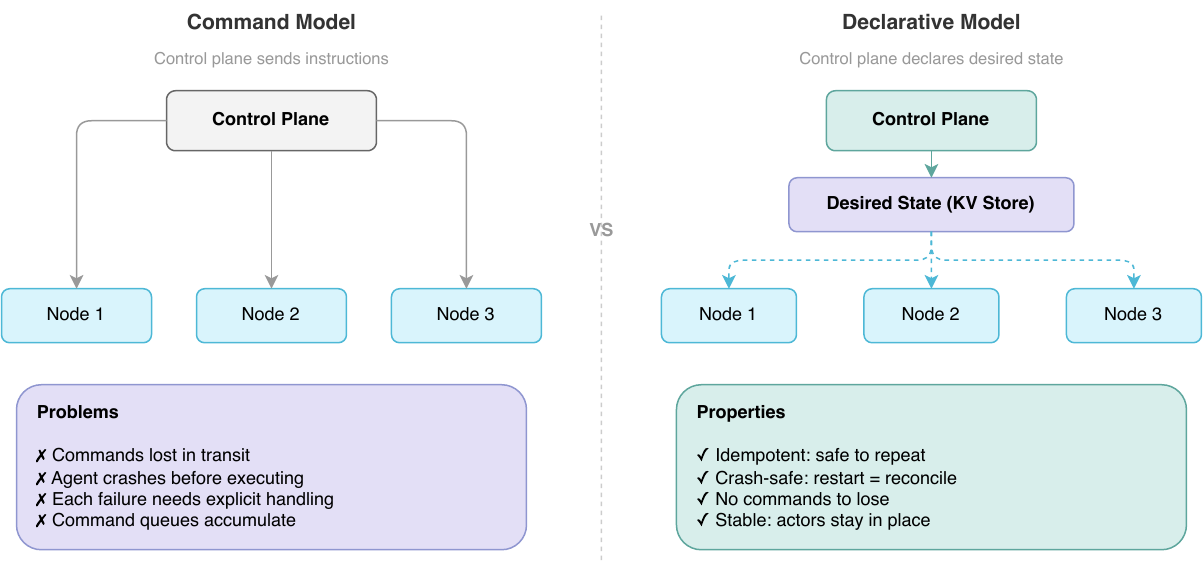}
    \caption{\textbf{Command model vs.\ declarative model.} In the command model (left), the control plane sends instructions directly to nodes, leading to problems such as lost commands and accumulated queues. In our declarative model (right), the control plane publishes desired state to a KV store, and nodes independently observe, diff, and reconcile, yielding idempotent, crash-safe operation.}
    \label{fig:infra_declarative}
\end{figure}

\subsubsection{Declarative Reconciliation Architecture}

\begin{figure}[!t]
    \centering
    \includegraphics[width=0.85\linewidth]{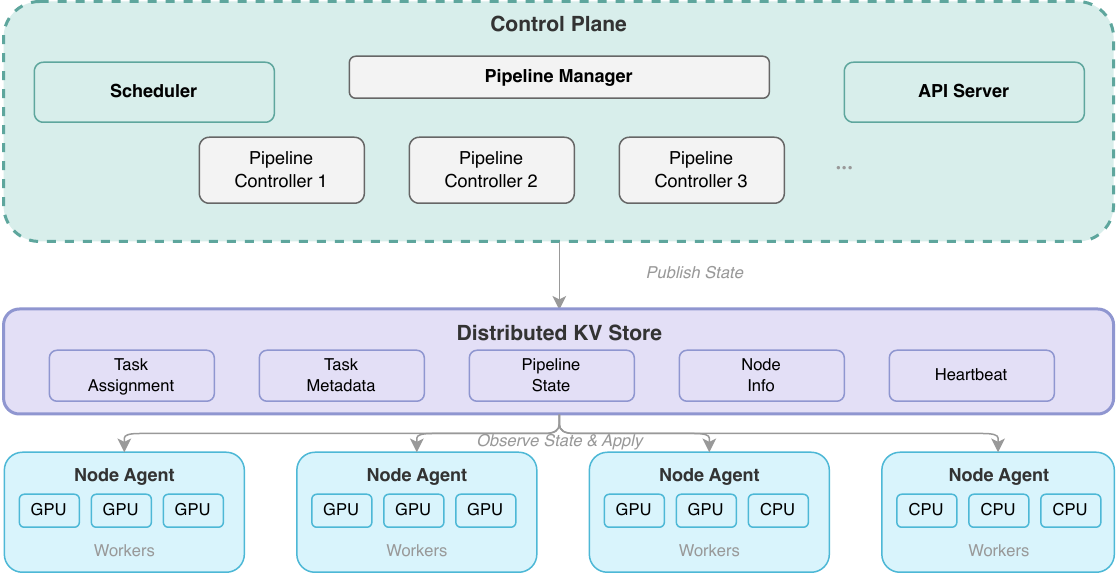}
    \caption{\textbf{Four-layer architecture of the data processing engine.} The control plane (scheduler and pipeline controllers) publishes desired state to a distributed KV store. Node agents observe state changes and reconcile local workers accordingly. Each layer has a single responsibility and can be independently restarted without affecting running pipelines.}
    \label{fig:infra_layers}
\end{figure}

The core architectural decision is a \emph{declarative} engine, drawing from the same principles behind Kubernetes. Instead of issuing imperative commands (\eg ``start actor X on node Y''), the control plane declares desired state in a distributed key-value store. Nodes independently converge local state toward the declared target through an \emph{observe $\rightarrow$ diff $\rightarrow$ reconcile} loop, replacing the entire command dispatch, acknowledgment, retry, and rollback complexity, as illustrated in Figure~\ref{fig:infra_declarative}.

This design is particularly well-suited to our workload: actors load large models onto GPUs with initialization costs measured in seconds to minutes. Once running, actors should remain in place to amortize that cost. The system optimizes for placement stability, not churn.

The engine is organized into four layers, each with a single responsibility (Figure~\ref{fig:infra_layers}):

\paragraph{Layer 1: Scheduler}
Reads demand, metrics, and capacity from the KV store and publishes placement assignments. GPU-bound stages are packed densely to minimize fragmentation; IO/CPU-bound stages are spread for throughput balance. Latency-critical pipelines are placed first; background pipelines expand into remaining capacity and contract on demand.

\paragraph{Layer 2: Pipeline Controllers}
One controller per pipeline declares demand for its stages, publishes workload metrics, wires the pipeline DAG, and manages lifecycle from code delivery to graceful shutdown. Controllers are fault-isolated: a controller crash affects only its pipeline, and restart re-reads state from the KV store and reconverges idempotently.

\paragraph{Layer 3: Node Agents}
One agent per worker node runs the core reconciliation loop (read desired state, compare with actual running processes, spawn or kill workers to close the gap) and reports status to the control plane periodically.

\paragraph{Layer 4: Workers}
Stateless, single-purpose, disposable processes. Each hosts one task instance, pulls work from a task queue, and writes output. Workers contain zero coordination logic and no awareness of the scheduler, peers, or topology. Scale-out and crash replacement use the same code path.

\subsubsection{Results}

The custom engine achieves several key operational improvements:

\begin{itemize}
    \item \textbf{GPU utilization above 95\%.} Priority-aware scheduling, hardware-aware bin-packing, and dynamic capacity reallocation eliminate GPU fragmentation. Background pipelines absorb every idle cycle that inference and latency-critical pipelines leave behind.
    \item \textbf{Highly available control plane.} Any transient failure (scheduler restart, KV failover, or network partition) does not affect running tasks. Workers continue processing, node agents continue reconciling, and when the control plane recovers, it reads current state and resumes without losing in-flight work.
    \item \textbf{Node failure detection in under 30 seconds.} Down from 5--10 minutes. On a cluster with constant inference preemption, this directly recovers GPU-hours that would otherwise be silently lost.
    \item \textbf{Linear scalability.} The control plane scales linearly with cluster size, supporting 5{,}000+ GPU nodes and 200K+ concurrent tasks without becoming the bottleneck, a capability that was structurally impossible with a centralized GCS.
    \item \textbf{Zero-downtime deployments.} Because all components are stateless and state lives in the KV store, any layer can be rolling-restarted without interrupting running pipelines.
\end{itemize}

% ============================================================
\section{Evaluation}
\label{sec:eval}

We evaluate \model{} through both objective automated metrics and subjective human evaluation, designed to assess identity-preserving avatar video generation across multiple perceptual dimensions. Our evaluation compares against four state-of-the-art systems spanning both avatar-specialized and general video generation models, using a diverse cross-scene benchmark that tests generalization beyond the training distribution.

\subsection{Benchmark Construction}

\paragraph{Test set} We construct a cross-scene evaluation benchmark comprising 70 test cases sourced from publicly available online videos, with a focus on talking-video scenarios. For each test case, we collect two video clips depicting the same individual in different scenes. One clip serves as the \emph{reference video} providing identity context, while the first frame and audio track of the other clip serve as the driving signals. This cross-scene setup is deliberately more challenging than same-identity-same-scene evaluation, as it requires the model to transfer identity information across visual contexts rather than simply reproducing the reference scene.

\begin{figure}[!t]
  \centering
  \includegraphics[width=\textwidth]{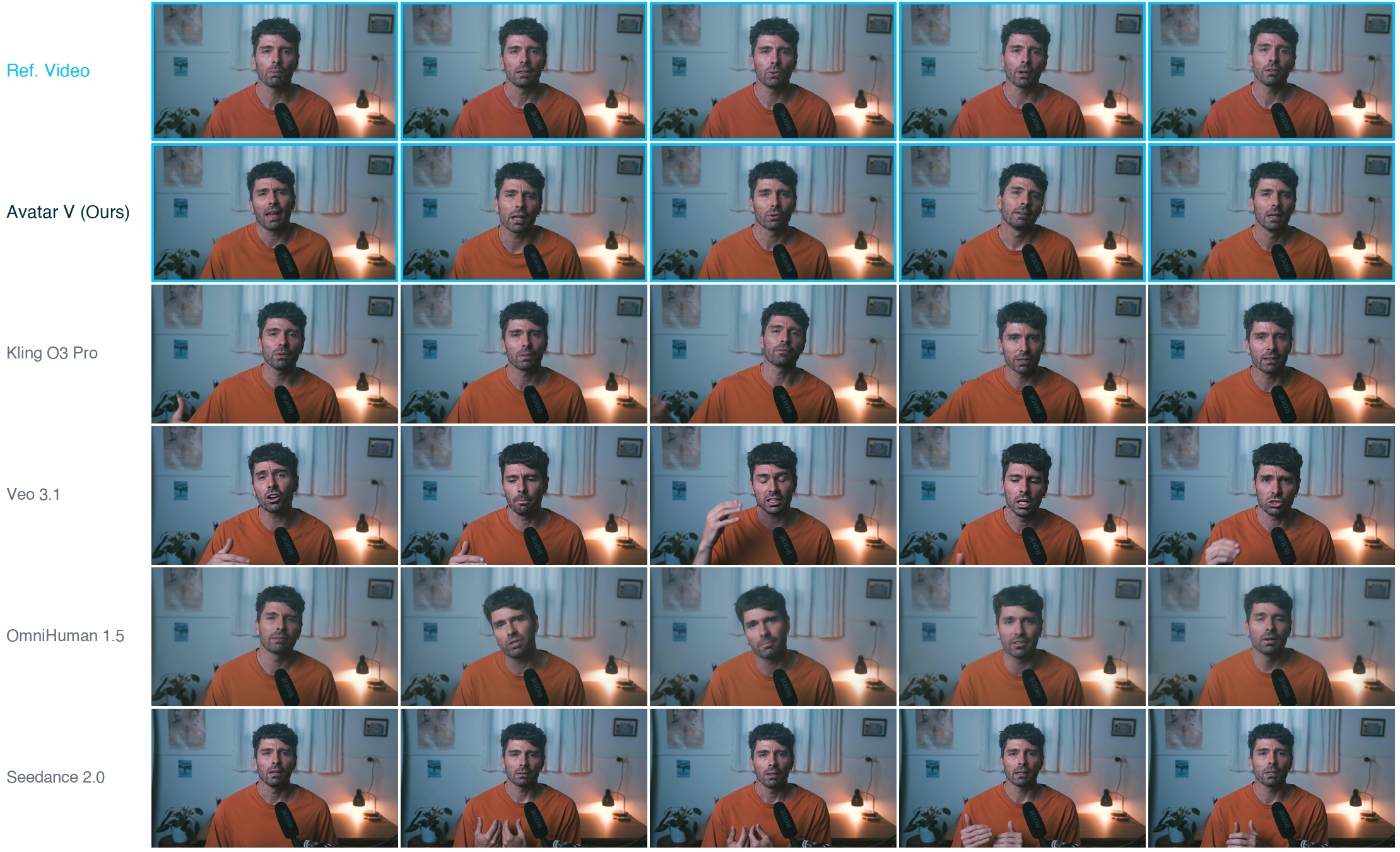}
  \caption{\textbf{Qualitative comparison: same-scene condition.} The reference video (cyan border, top row) provides identity context. All five methods generate from the same driving audio and scene image. \model{} produces the most faithful identity and natural motion.\protect\footnotemark}
  \label{fig:eval_qual_same}
\end{figure}

\begin{figure}[!t]
  \centering
  \includegraphics[width=\textwidth]{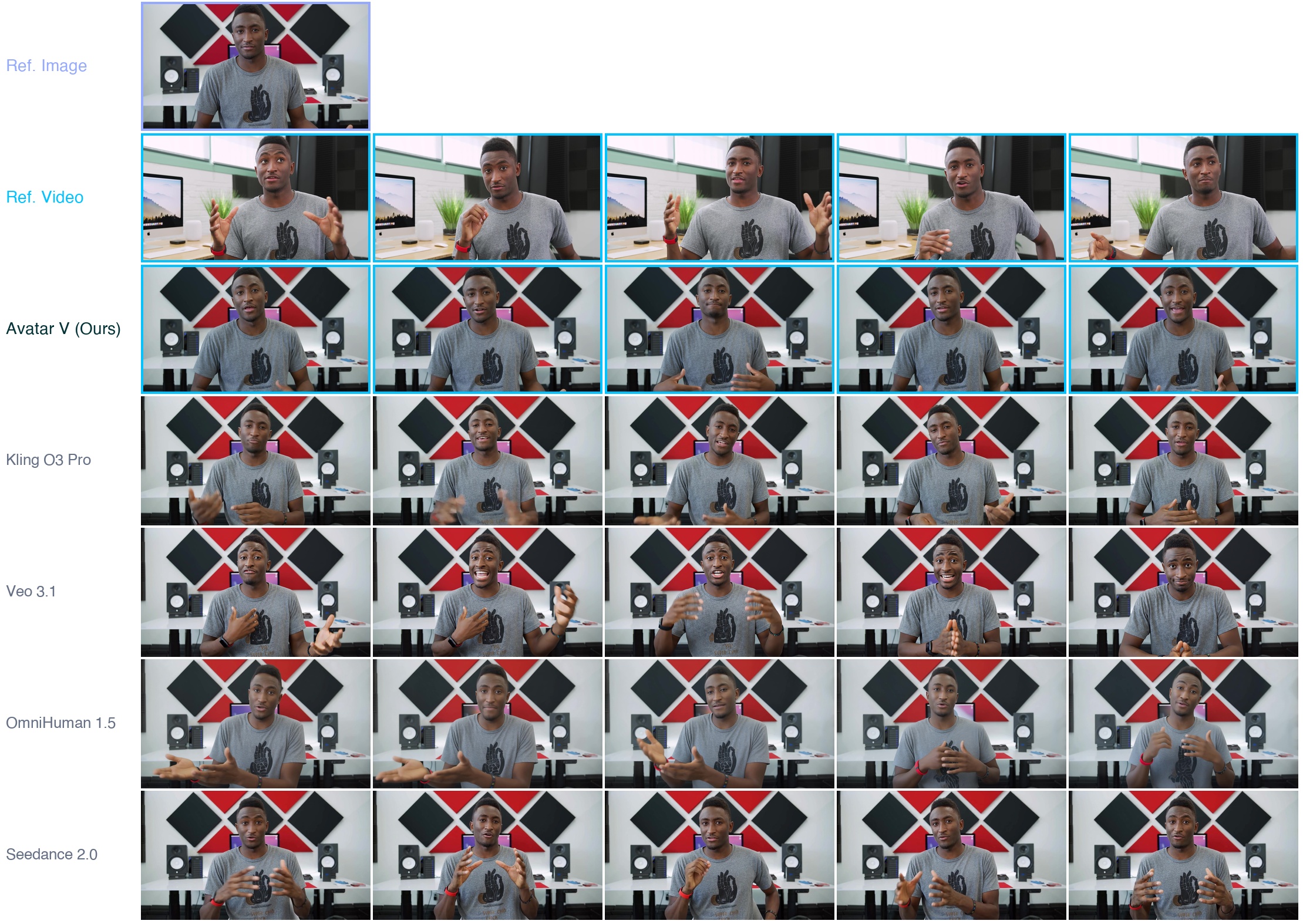}
  \caption{\textbf{Qualitative comparison: cross-scene condition.} The driving scene image (purple border, top-left) differs from the reference video scene, requiring cross-context identity transfer.}
  \label{fig:eval_qual_cross}
\end{figure}

\begin{figure}[!t]
  \centering
  \includegraphics[width=\textwidth]{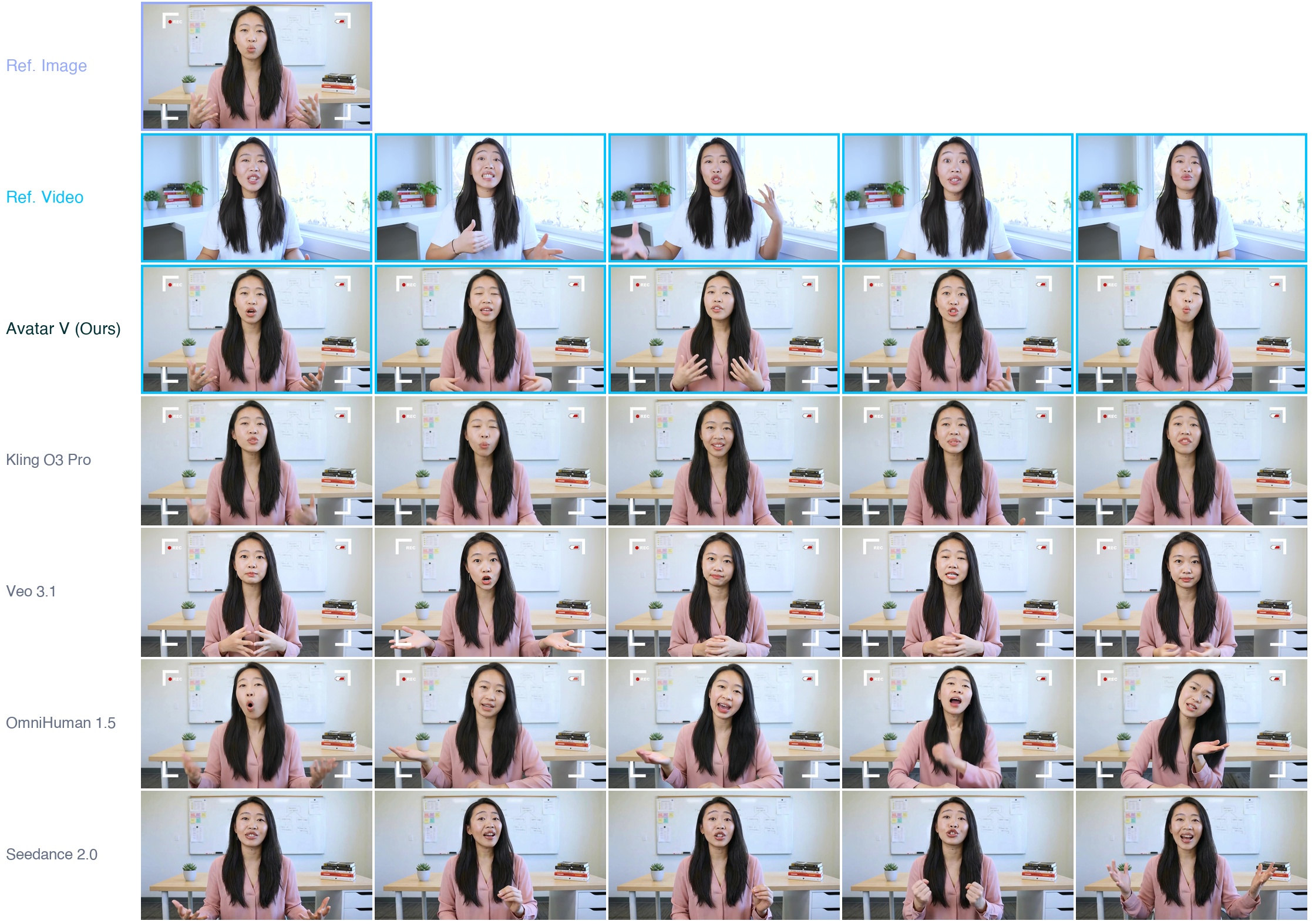}
  \caption{\textbf{Qualitative comparison: generated-scene condition.} The scene image (purple border, top-left) is produced by the Identity-Preserving Image Engine, representing the fully automated production pipeline.}
  \label{fig:eval_qual_gen}
\end{figure}

\paragraph{Scene conditions} To evaluate robustness under varying scene configurations, we test three scene generation modes: (1)~\emph{Same-scene}, where the target scene image is drawn from the same scene as the reference video, providing an upper-bound on scene familiarity; (2)~\emph{Cross-scene}, where the target scene image comes from a different real video of the same individual, testing cross-context generalization; and (3)~\emph{Generated-scene}, where the scene image is produced by our Identity-Preserving Image Engine (Section~\ref{sec:model}), representing the fully automated production pipeline. Figures~\ref{fig:eval_qual_same},~\ref{fig:eval_qual_cross}, and~\ref{fig:eval_qual_gen} show representative examples from each condition.

\footnotetext{All videos shown in this report are for research demonstration purposes only. HeyGen's platform enforces consent verification for all digital twin creation.}

In the cross-scene condition (Figure~\ref{fig:eval_qual_cross}), the target scene image is drawn from a different video of the same individual, requiring the model to disentangle identity from scene-specific details. Methods that rely on scene cues rather than genuine identity features exhibit noticeable quality degradation, while \model{} maintains faithful identity transfer across visually distinct contexts.

The generated-scene condition (Figure~\ref{fig:eval_qual_gen}) represents the fully automated production pipeline, where the scene image is synthesized by the Identity-Preserving Image Engine. This is the most challenging setting as it combines identity transfer with a novel, unseen background. \model{} produces temporally coherent outputs with natural expressions, while several competing methods struggle with identity consistency or introduce visible artifacts.

\paragraph{Competing methods} We compare against four systems: Kling O3 Pro, Veo 3.1, OmniHuman 1.5, and Seedance 2.0, accessed through their latest publicly available versions at the time of evaluation. Some competitors generate audio that does not match the original speech; in these cases, we evaluate the visual output independently.

\subsection{Objective Evaluation}

We compute four automated metrics, each targeting a distinct aspect of generation quality. All per-frame metrics are sampled at 2\,fps and computed as 10\% trimmed means (removing the top and bottom 10\% of frames) for robustness.

\paragraph{SyncNet score} We measure audio-visual synchronization at the video level using SyncNet~\citep{chung2017syncnet}. The model detects face tracks across frames, computes audio-visual embeddings for each track segment, and reports the synchronization confidence (Sync-C, higher is better) and minimum distance (Sync-D, lower is better).

\paragraph{Face similarity} Identity preservation is measured as the cosine similarity between ArcFace~\citep{deng2019arcface} embeddings of detected faces in each generated frame and the reference image.

\paragraph{Q-Align IQA} We assess overall frame-level perceptual quality using Q-Align~\citep{wu2023qalign}, a vision-language model that produces quality scores calibrated to human mean opinion scores.

\subsubsection{Results}

\begin{table}[!t]
\caption{\textbf{Objective metrics comparison} on the 36 matched test cases where all five methods produced valid outputs. Per-frame metrics (Face Sim, Q-Align) are computed as 10\% trimmed means for robustness. Higher is better for all metrics ($\uparrow$) except Sync-D ($\downarrow$). Best results are in \textbf{bold}, second best \underline{underlined}.}
\label{tab:obj_results}
\centering
\small
\begin{tabular}{@{}lcccc@{}}
\toprule
\textbf{Method} & \textbf{Sync-C$\uparrow$} & \textbf{Sync-D$\downarrow$} & \textbf{Face Sim$\uparrow$} & \textbf{Q-Align$\uparrow$} \\
\midrule
Ground Truth & 7.93 & 6.76 & 0.861 & 4.75 \\
\midrule
Kling O3 Pro (2026) & 5.16 & 10.07 & \underline{0.838} & 4.80 \\
Veo 3.1 (2025) & 8.05 & 7.28 & 0.714 & \textbf{4.95} \\
OmniHuman 1.5 (2025) & 7.53 & 8.25 & 0.732 & 4.70 \\
Seedance 2.0 (2026) & \underline{8.86} & \underline{6.99} & 0.823 & \underline{4.85} \\
\midrule
\model{} (Ours) & \textbf{8.97} & \textbf{6.75} & \textbf{0.840} & \underline{4.85} \\
\bottomrule
\end{tabular}
\end{table}

\model{} achieves the strongest overall performance across all four automated metrics. On lip synchronization, \model{} achieves the highest SyncNet confidence (8.97) and the lowest Sync-D (6.75), surpassing even ground truth recordings (7.93 / 6.76) and demonstrating superior audio-visual alignment. On identity preservation, \model{} achieves the highest Face Similarity (0.840) among all methods, closely approaching ground truth (0.861) and substantially outperforming Veo 3.1 (0.714) and OmniHuman 1.5 (0.732). On Q-Align perceptual quality, \model{} ties with Seedance 2.0 for second place (4.85). Veo 3.1 achieves the highest Q-Align score (4.95) but at the cost of severely degraded identity preservation (Face Sim = 0.714); we also observe that Veo 3.1 outputs exhibit noticeable over-sharpening, which can inflate perceptual quality scores without corresponding subjective improvement. This analysis reveals a key trade-off among competing methods: systems optimized for visual quality metrics may sacrifice identity fidelity or resort to aggressive post-processing. \model{} uniquely maintains top-tier performance across identity, synchronization, and quality axes simultaneously.

\subsection{Subjective Evaluation}

All subjective evaluations are conducted by trained human annotators following the annotation system described in Section~\ref{sec:data}.

\subsubsection{Mean Opinion Score (MOS)}

Each generated video is independently rated on a 5-point Likert scale (1 = very poor, 5 = excellent) across six perceptual dimensions: identity consistency, lip-sync accuracy, motion naturalness, motion consistency, artifact control, and visual quality. Each video is rated by at least two annotators, blinded to model identity, in randomized order. The final score is the arithmetic mean.

\begin{table}[!t]
\caption{\textbf{MOS comparison} (5-point Likert scale). Higher is better. Best results are in \textbf{bold}, second best \underline{underlined}.}
\label{tab:mos_results}
\centering
\small
\begin{tabular}{@{}lcccccc@{}}
\toprule
\textbf{Method} & \textbf{Identity$\uparrow$} & \textbf{Lip Sync$\uparrow$} & \textbf{Motion Nat.$\uparrow$} & \textbf{Motion Con.$\uparrow$} & \textbf{Artifacts$\uparrow$} & \textbf{Visual$\uparrow$} \\
\midrule
Kling O3 Pro (2026) & 4.18 & 4.40 & \underline{4.21} & 4.12 & 4.19 & 4.45 \\
Veo 3.1 (2025) & 4.34 & 4.62 & 3.88 & 4.05 & \underline{4.66} & \underline{4.76} \\
OmniHuman 1.5 (2025) & 4.70 & 4.04 & 3.59 & 3.87 & 3.89 & 3.81 \\
Seedance 2.0 (2026) & \underline{4.84} & \underline{4.64} & 4.13 & \underline{4.44} & 4.61 & 4.17 \\
\midrule
\model{} (Ours) & \textbf{4.98} & \textbf{4.69} & \textbf{4.48} & \textbf{4.57} & \textbf{4.75} & \textbf{4.78} \\
\bottomrule
\end{tabular}
\end{table}

\begin{figure}[!t]
  \centering
  \includegraphics[width=0.55\textwidth]{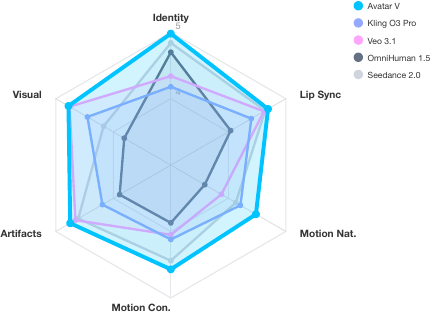}
  \caption{\textbf{MOS radar chart.} \model{} (cyan) consistently achieves the highest human ratings across all six perceptual dimensions, demonstrating balanced excellence rather than specialization in a single aspect.}
  \label{fig:eval_radar_mos}
\end{figure}

\model{} achieves the highest MOS scores on all six dimensions (Table~\ref{tab:mos_results}, Figure~\ref{fig:eval_radar_mos}). On identity consistency, \model{} scores 4.98 out of 5, nearly perfect and substantially higher than all competitors. The advantage is particularly pronounced on motion naturalness (4.48 vs. second-best 4.21) and motion consistency (4.57 vs. 4.44), reflecting the effectiveness of the dedicated motion representation in capturing individual-specific behavioral patterns. Unlike competing methods that excel on one or two aspects while underperforming on others (e.g., Veo 3.1 scores high on visual quality but low on motion naturalness; OmniHuman 1.5 preserves identity but produces artifacts), \model{} maintains uniformly high quality across all dimensions.

\subsubsection{Pairwise Win Rate}

For pairwise relative evaluation, we present annotators with side-by-side video pairs: one from \model{} and one from a competing method, both generated from identical inputs. Annotators select which video they prefer as an overall quality judgment. Each pair is rated by three annotators, and the final winner is determined by majority vote.

\begin{table}[!t]
\caption{\textbf{Pairwise win rate.} Each row shows the percentage of test cases where \model{} is preferred (Win) or the competitor is preferred (Lose), determined by majority vote of three annotators. The number of evaluated cases varies across competitors because certain commercial APIs reject inputs containing celebrity likenesses due to portrait rights restrictions.}
\label{tab:winrate_results}
\centering
\small
\begin{tabular}{@{}lccc@{}}
\toprule
\textbf{Competitor} & \textbf{Win$\uparrow$} & \textbf{Lose$\downarrow$} & \textbf{\#Cases} \\
\midrule
Kling O3 Pro & 69.6\% & 30.4\% & 69 \\
Seedance 2.0 & 68.9\% & 31.1\% & 45 \\
Veo 3.1 & 72.5\% & 27.5\% & 40 \\
OmniHuman 1.5 & 85.7\% & 14.3\% & 70 \\
\bottomrule
\end{tabular}
\end{table}

\begin{figure}[!t]
  \centering
  \includegraphics[width=0.90\textwidth]{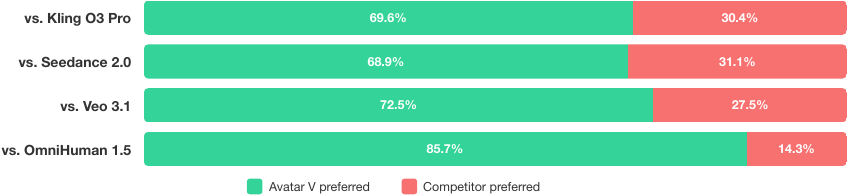}
  \caption{\textbf{Pairwise win rate.} Each bar shows the percentage of cases where \model{} is preferred (green) or the competitor is preferred (red) relative to each competitor.}
  \label{fig:eval_winrate}
\end{figure}

As shown in Table~\ref{tab:winrate_results}, \model{} is consistently preferred across all four competitors, achieving win rates ranging from 68.9\% to 85.7\%. The advantage is most pronounced against OmniHuman 1.5 (85.7\%) and Veo 3.1 (72.5\%), while \model{} maintains a clear lead against Kling O3 Pro (69.6\%) and Seedance 2.0 (68.9\%).

Qualitative analysis of annotator feedback reveals two consistent strengths of \model{}: (1)~\emph{generation stability}, where \model{} rarely produces visual artifacts across diverse test cases, and (2)~\emph{holistic identity consistency}, where both the subject's appearance and behavioral patterns and talking style closely match the reference video, yielding high perceived similarity under visual inspection.

In contrast, competing methods exhibit several recurring failure modes: excessive skin smoothing and over-enhancement that reduces perceived realism; poor behavioral consistency, particularly in motion dynamics and gestural patterns that deviate from the reference; and unstable generation quality with various visual artifacts across frames.

\subsubsection{Avatar Turing Test}

To evaluate perceptual realism, we conduct an Avatar Turing Test in which annotators are shown pairs of videos consisting of an Avatar V generation and its corresponding ground-truth recording, and are asked to determine which video is real. In this setup, a system approaching 50\% identification accuracy would be considered perceptually indistinguishable from real footage.

\begin{table}[!t]
\caption{\textbf{Avatar Turing Test results.} Three annotators each evaluated 18 pairs of \model{} outputs and ground truth videos, yielding 54 total judgments. Lower identification accuracy indicates more realistic generation.}
\label{tab:turing}
\centering
\small
\begin{tabular}{@{}lc@{}}
\toprule
\textbf{Metric} & \textbf{Value} \\
\midrule
Real identification accuracy & 77.8\% \\
Fooled rate (\model{} mistaken as real) & 22.2\% \\
Chance level (random guessing) & 50.0\% \\
\midrule
Cases fooling $\geq$1 annotator & 11/18 (61.1\%) \\
\bottomrule
\end{tabular}
\end{table}

Across all judgments, annotators correctly selected the real video 77.8\% of the time, indicating that the generated videos remain distinguishable overall. However, the results also show strong realism: in 61.1\% of test cases, at least one of three annotators identified the generated video as the real one. This suggests that Avatar V is able to produce highly realistic talking-avatar videos that can frequently deceive trained evaluators on a case-by-case basis, even though a measurable gap from full perceptual indistinguishability still remains.

% ============================================================
\section{Related Work}
\label{sec:related_work}

\subsection{Video Diffusion Models}
\label{sec:video_diffusion}

Diffusion-based video generation has advanced rapidly since Sora~\cite{brooks2024sora} demonstrated that Diffusion Transformers (DiT)~\cite{peebles2023scalable} can generate high-fidelity minute-long videos.
A wave of large-scale models followed, including CogVideoX~\cite{cogvideox2024} with joint spatiotemporal 3D VAE compression, HunyuanVideo~\cite{hunyuanvideo2024} as an open-source 13B model competitive with closed-source systems, Wan~\cite{wan2025} validating scaling laws at both 14B and 1.3B scales, Movie Gen~\cite{moviegen2024} unifying video and audio generation at 30B parameters, Kling~\cite{kling2024}, Lumiere~\cite{lumiere2024}, Step-Video~\cite{stepvideo2025}, and SkyReels-V2~\cite{skyreelsv2_2025}.
Architecturally, DiT has largely replaced U-Net as the predominant backbone~\cite{gentron2024,latte2025}, while Flow Matching~\cite{lipman2022flow,pyramidalflow2025} has emerged as a practical alternative to DDPM~\cite{ho2020denoising} training objectives. Open-Sora~\cite{opensora2024,opensora2_2025} demonstrated commercial-grade quality at limited budget, and Cosmos~\cite{cosmos2025} positioned video generation as a world foundation model platform.
For controllable generation, Stable Video Diffusion~\cite{blattmann2023stable} studied transfer from image to video diffusion, DynamiCrafter~\cite{dynamicrafter2024} animates open-domain images, and AnimateDiff~\cite{animatediff2024} proposed plug-and-play motion modules compatible with personalized text-to-image models.
\model{} builds upon this DiT and flow matching foundation but diverges in two key ways: it introduces an asymmetric attention mechanism tailored for identity-conditioned generation, and it employs a progressive training pipeline that extends beyond standard text-to-video objectives to audio-driven, personality-preserving synthesis.

\subsection{Portrait Video Generation}
\label{sec:portrait_video}

\paragraph{Audio-driven talking head generation}
Early work such as SadTalker~\cite{sadtalker2023} mapped audio to 3DMM motion coefficients for decoupled head motion and expression synthesis.
Diffusion-based methods brought significant quality improvements: EMO~\cite{emo2024} pioneered direct audio-to-video generation without intermediate representations; Hallo~\cite{hallo2024} introduced hierarchical audio conditioning for separate control of lip, expression, and pose, with Hallo2~\cite{hallo2_2024} extending to longer durations and higher resolutions, and Hallo3~\cite{cui2025hallo3} and Hallo4~\cite{cui2025hallo4} further advancing controllability and visual quality; EchoMimic~\cite{echomimic2024} added editable landmark control, followed by EchoMimicV3~\cite{meng2026echomimicv3} which unified multiple generation tasks in a single architecture; V-Express~\cite{vexpress2024} addressed weak-signal suppression in multi-condition models; and VASA-1~\cite{vasa1_2024} achieved real-time generation via a latent-space DiT.
More recent works have tackled multi-condition orchestration: OmniHuman~\cite{lin2025omnihuman,lin2025omnihuman15} proposed a one-stage conditioned human animation framework that scales to high-quality generation with cognitive simulation, OmniAvatar~\cite{gan2025omniavatar} proposed a unified framework for diverse avatar tasks, HuMo~\cite{chen2025humo} introduced collaborative multi-modal conditioning for synchronized body and face generation, and StableAvatar~\cite{tu2025stableavatar} addressed infinite-length generation with identity consistency.
A common limitation of these methods is their reliance on single-image references, which constrains identity information to one viewpoint and expression. \model{} overcomes this by conditioning on full video references through Sparse Reference Attention, extracting rich static and dynamic identity cues without per-identity fine-tuning.

\paragraph{Single-image portrait animation}
LivePortrait~\cite{liveportrait2024} achieves real-time inference through implicit keypoints with stitching and retargeting modules.
AniPortrait~\cite{aniportrait2024} combines facial landmarks with audio for controllable portrait animation.
X-Portrait~\cite{xportrait2024} handles large head movements via hierarchical motion attention, and MuseTalk~\cite{musetalk2024} enables efficient real-time lip synchronization through latent-space inpainting.
While these single-image approaches are efficient, they fundamentally lack the dynamic identity signals (talking rhythm, habitual expressions) that video references provide.

\paragraph{Video-reference-based approaches}
Several recent works explore conditioning generation on video references rather than single images. WanAnimate~\cite{cheng2025wananimate} extends Wan with video-guided motion transfer, SlotID~\cite{lai2026slotid} uses slot attention for identity-disentangled control, and Seedance 2.0~\cite{seedance2_02025} incorporates full-clip reference conditioning.
However, these approaches either compress references through bottleneck encoders that discard fine-grained details, or concatenate all reference tokens incurring quadratic attention cost. \model{} addresses both issues through its asymmetric sparse attention design, where generation tokens attend to reference tokens while reference tokens only self-attend, achieving linear complexity in reference length.

\subsection{Human Body Video Generation}
\label{sec:human_body_video}

\paragraph{Pose-guided human animation}
Animate Anyone~\cite{animateanyone2024} established the paradigm of ReferenceNet combined with Pose Guider and Temporal Attention for single-image pose-driven animation, while MagicAnimate~\cite{magicanimate2024} introduced an appearance encoder with video fusion for temporal consistency.
Subsequent works enriched the guidance signal: Champ~\cite{champ2024} incorporated SMPL-derived 3D conditions (depth, normals, semantics), MimicMotion~\cite{mimicmotion2024} introduced confidence-aware pose guidance for stable long-sequence generation, and UniAnimate~\cite{unianimate2024} unified reference, pose, and video in a shared latent space for minute-length synthesis. MIMO~\cite{mimo2024} extended to scene-controllable multi-character generation via spatial decomposition.

\paragraph{Identity preservation}
Maintaining character consistency remains a core challenge. IP-Adapter~\cite{ipadapter2024} achieves image-text prompt compatibility through decoupled cross-attention, InstantID~\cite{instantid2024} combines identity embeddings with landmark guidance for zero-shot preservation, and PhotoMaker~\cite{photomaker2024} supports multi-reference identity fusion via stacked embeddings. These techniques provide critical foundations for reference-based conditioning in human video generation. Unlike bottleneck-based identity encoders, \model{} retains the full visual richness of reference tokens through its video-reference attention mechanism, avoiding the information loss inherent in fixed-size identity embeddings.

\subsection{Training Efficiency and Alignment}
\label{sec:training_efficiency}

\paragraph{Diffusion distillation}
Reducing the inference cost of diffusion models has been widely studied. Progressive distillation~\cite{salimans2022progressive} halves the number of sampling steps iteratively, while consistency models~\cite{song2023consistency} learn to map any noisy sample directly to the clean output. Distribution Matching Distillation (DMD)~\cite{yin2024dmd} introduces a regression loss combined with an adversarial objective for few-step generation. Classifier-free guidance (CFG) distillation~\cite{meng2023distillation} internalizes the conditional and unconditional score combination, eliminating the need for multiple forward passes. \model{} combines CFG distillation with DMD in a two-phase pipeline, achieving over $10\times$ inference acceleration while maintaining generation quality.

\paragraph{Reinforcement learning for generative models}
Aligning generative models with human preferences has drawn increasing attention. DPO~\cite{rafailov2023dpo} provides a reference-model-free approach to preference optimization, and Diffusion-DPO~\cite{wallace2024diffusiondpo} adapts it to diffusion training. RLHF for diffusion~\cite{black2024rlhfdiffusion} directly optimizes reward functions through policy gradients. More recently, GRPO~\cite{shao2024grpo}, originally proposed for language models, has been adapted to visual generation: DanceGRPO~\cite{huang2025dancegrpo} and FlowGRPO~\cite{zhang2025flowgrpo} apply group-relative advantage estimation to video diffusion, demonstrating the feasibility of multi-reward RL for visual content. \model{} extends this line of work with identity, motion, and visual quality reward functions tailored to avatar generation, combined with DPO for complementary preference alignment.

% ============================================================
\section{Ethics and Safety}
\label{sec:ethics}

Avatar generation raises important considerations around consent and content safety. Our production platform addresses these through two mechanisms. First, creating a custom avatar requires explicit verification from the individual being represented; the depicted individual retains the right to request removal of their likeness at any time. Second, all content uploaded to or generated by the platform passes through a two-stage moderation pipeline combining automated review powered by machine learning with manual review by human moderators, covering categories including but not limited to fraud, harassment, child safety, misinformation, and intellectual property infringement. Violations may result in content removal, account suspension, or reporting to legal authorities. The full policy is available at \url{https://www.heygen.com/moderation-policy}.

% ============================================================
\section{Conclusion}
\label{sec:conclusion}

We have presented \model{}, a system for generating high-fidelity talking avatar videos from short video references. The central insight is to formulate identity conditioning as a video-reference conditioning problem: by letting the model attend directly to the full token sequence of a reference video through Sparse Reference Attention, \model{} captures both static appearance features (facial geometry, skin texture, accessories) and dynamic behavioral patterns (talking rhythm, habitual expressions, gestural tendencies) without per-identity fine-tuning or information-lossy bottleneck encoders.

Around this core mechanism, we contribute a suite of techniques spanning model design, data, training, and deployment: a dedicated motion representation stream that creates closed-loop supervision for person-specific talking style through joint generation and conditioning; an identity-aware super-resolution refiner with sparse temporal attention that recovers fine facial details at high resolution; an LLM-based voice cloning engine that reproduces the target speaker's vocal identity from a short audio sample; a scalable data curation pipeline processing 50M+ raw videos into 100M+ pretraining clips and 10M+ avatar fine-tuning clips with cross-clip identity connectivity; a five-stage progressive training pipeline incorporating text-to-video pretraining, audio-to-video pretraining, personality SFT, two-phase distillation for over $10\times$ inference acceleration, and reinforcement learning from human feedback; and a comprehensive inference optimization stack deployed across thousands of GPUs under a unified multi-cloud infrastructure.

Experiments on our cross-scene benchmark demonstrate that \model{} achieves state-of-the-art performance across all evaluated dimensions, including identity preservation, lip synchronization, expression naturalness, motion quality, and visual fidelity, as measured by both automated metrics and comprehensive human evaluation.

% ============================================================
\section*{Contributions and Acknowledgments}
\label{sec:contributions}
\addcontentsline{toc}{section}{Contributions and Acknowledgments}

All contributors are listed in alphabetical order by first name.

\vspace{1em}

\begin{multicols}{3}
\raggedright
\noindent
Benjamin Liang\\
Ce Chen\\
Desmond Lin\\
Ivan Somov\\
Jiajun Zhao\\
Jiewei Yuan\\
Jingfeng Zhang\\
Junhao Huang\\
Nik Nolte\\
Pedram Haqiqi\\
Penghan Wang\\
Rong Yan\\
Rui Zhang\\
Sam Prokopchuk\\
Sivan Wang\\
Viktor Goriachko\\
Yi Ren\\
Yuanming Li\\
Yutao Chen\\
Zhenhui Ye\\
Zhibin Hong\\
Zilong Nie\\
Zujin Guo\\
\end{multicols}

{\small
\bibliographystyle{plainnat}
\bibliography{main}
}

\end{document}